\documentclass{article}

\usepackage[preprint]{neurips_2026}
\makeatletter
\renewcommand{\@noticestring}{}
\makeatother

\usepackage[utf8]{inputenc}
\usepackage[T1]{fontenc}
\usepackage{hyperref}
\usepackage{url}
\usepackage{booktabs}
\usepackage{amsfonts}
\usepackage{nicefrac}
\usepackage{microtype}
\usepackage{xcolor}

\usepackage{graphicx}
\usepackage{subcaption}
\usepackage{array}
\usepackage{tabularx}
\usepackage[section]{placeins}
\usepackage{amsmath}
\usepackage{amssymb}
\usepackage{mathtools}
\usepackage{amsthm}
\usepackage[capitalize,noabbrev]{cleveref}

\let\sectionwithoutfloatbarrier\section

\newcolumntype{Y}{>{\raggedright\arraybackslash}X}
\newcounter{pipelinefigure}
\renewcommand{\thepipelinefigure}{C.\arabic{pipelinefigure}}
\newcounter{sensitivityfigure}
\renewcommand{\thesensitivityfigure}{J.\arabic{sensitivityfigure}}
\newcommand{\promptblock}[3]{%
  \subsection*{\texttt{#1}}
  \begingroup
  \small
  \setlength{\tabcolsep}{6pt}
  \renewcommand{\arraystretch}{1.15}
  \noindent\begin{tabularx}{\linewidth}{@{}>{\bfseries}p{0.17\linewidth}Y@{}}
  System & #2 \\
  User & #3 \\
  \end{tabularx}
  \par\endgroup\medskip
}

\theoremstyle{plain}

\theoremstyle{definition}

\theoremstyle{remark}

\title{Detecting Hidden Chain-of-Thought in Large Language Models with Linguistic, Behavioral, and Mechanistic Indicators}
\hypersetup{
  pdftitle={Detecting Hidden Chain-of-Thought in Large Language Models with Linguistic, Behavioral, and Mechanistic Indicators},
  pdfauthor={Armaan Singh, Ryan Trinh Le, Jasmine Kaur, Abdullah Sultan, Edward Lue Chee Lip, Kiran Nijjer, Adnan Ahmed, Vasu Sharma},
  pdfsubject={Machine Learning Research Paper}
}

\author{
\bfseries Armaan Singh$^{1}$, Ryan Trinh Le$^{1}$, Jasmine Kaur$^{1}$, Abdullah Sultan$^{1}$,\\
\bfseries Edward Lue Chee Lip, Kiran Nijjer$^{1}$, Adnan Ahmed$^{2}$, Vasu Sharma$^{3}$
}

\begin{document}

\maketitle

\begingroup
\renewcommand{\thefootnote}{}
\footnotetext{Code: \url{https://github.com/a4maan/detecting-hct}}
\footnotetext{Correspondence to: Armaan Singh $\langle$\href{mailto:armaan.singh@griffinschool.org}{\texttt{armaan.singh@griffinschool.org}}$\rangle$.}
\footnotetext{$^{1}$Stanford University, $^{2}$York University, $^{3}$Facebook}
\endgroup

\begin{abstract}
Large language models often answer complex reasoning questions without revealing intermediate steps, raising whether they reason latently or complete patterns. We propose the Hidden CoT Detection Score (HCDS), a comparative behavioral and mechanistic signal measuring whether neutral-prompt behavior aligns more closely with explicit CoT or explicit no-CoT. Here, \emph{hidden CoT} operationally denotes this neutral-prompt CoT-like alignment; HCDS does not directly observe or prove an unexposed reasoning trace. On GSM8K, HCDS is significantly positive for both Qwen3-4B variants (Thinking $+1.87$, $p = 1.2 \times 10^{-7}$; Instruct $+1.41$, $p = 1.9 \times 10^{-4}$), replicates across a different inference stack and quantization within $0.08$ ($+1.80$ and $+1.45$), and is not significantly positive in seven of eight length-adjusted calibration-control cells. The unadjusted score produces large positive scores on single-step arithmetic and numeric factual lookup. The variants also respond differently to no-CoT instructions: Instruct complies from the prompt alone, whereas Thinking continues reasoning and requires intervention. These findings show stronger, less prompt-conditional CoT-like behavior in the reasoning-tuned model, consistent with but not proof of latent reasoning. HCDS thus investigates latent reasoning without relying on models' self-reported traces.
\end{abstract}

\sectionwithoutfloatbarrier{Introduction}

Explicit Chain-of-Thought (CoT) prompting can substantially improve LLM performance on complex reasoning tasks while producing steps that appear to expose the model's reasoning~\citep{wei_chain--thought_2023}. This raises whether models perform similar multi-step computation internally when not explicitly asked to reason. The distinction matters for interpretability: reasoning-like text need not faithfully reflect the computations producing an answer~\citep{turpin_language_2023}, while genuine intermediate computation may occur without visible traces.

We study latent reasoning through behavioral identification. HCDS measures whether neutral-prompt behavior more closely resembles explicit CoT or explicit no-CoT behavior; this alignment is evidence consistent with latent reasoning, not direct observation of an internal trace. It compares three prompt conditions through six linguistic, behavioral, and mechanistic features: mean token entropy, entropy slope, latency per output token, paraphrase consistency, perturbation sensitivity, and the accuracy cost of suppressing steps identified by gradient$\times$activation attribution. Within each backend and $(\mathrm{model},\mathrm{dataset})$ group, raw features are first residualised on log generation length and the residuals are then z-scored before Euclidean distances are computed. HCDS subtracts the neutral prompt's distance to the CoT pole from its distance to the no-CoT pole.

Two design choices are critical. First, reasoning-tuned models may ignore no-CoT instructions: in a separate un-intervened GSM8K $n=500$ diagnostic with a 1024-token cap, Qwen3-4B Thinking produced a mean of 575 output tokens (median 490.5) when instructed not to reason. For the primary $n=50$ analysis, we force-close its reasoning block to obtain an answer-only comparison pole with median output of 6 tokens. Second, because most HCDS features correlate with output length, we residualise each on log generation length; length serves only for adjustment and diagnostics, never as a feature.

We evaluate HCDS on Qwen3-4B Instruct and Thinking using GSM8K and StrategyQA. On GSM8K, HCDS is significantly positive under neutral prompting for both variants (Table~\ref{tab:main-hcds}), indicating that neutral behavior is closer to explicit CoT than to explicit no-CoT. StrategyQA is reported separately (Appendix~\ref{app:strategyqa}) because its boolean answer format leaves the no-CoT pole above chance, weakening the resulting contrast. The variants also differ in instruction following: Instruct complies with no-CoT prompting, whereas Thinking does not.

Together, these results provide a framework combining behavioral alignment with mechanistic interventions to study reasoning not explicitly exposed in model outputs~\citep{chen_reasoning_2025} and test whether identified internal computations influence those outputs~\citep{meng_locating_2023}.

\section{Related Work}

\citet{turpin_language_2023} and \citet{chen_reasoning_2025} show that
chain-of-thought explanations can be unfaithful and need not reveal the
computations responsible for an answer. Rather than ask whether an explicit
rationale is faithful, we use behavioral and mechanistic signals to ask whether
latent reasoning occurs at all under neutral prompting.

\citet{lin_implicit_2025} find that implicit reasoning can arise through brittle
shortcut learning, while \citet{he_reasoning_2026} identify steerable internal
features associated with latent reasoning. We instead detect inference-time use
of latent reasoning rather than study its emergence during training or induction
through steering.

The closest work to our paraphrase-consistency feature is
\citet{shi_finding_2026}: their PPCV framework uses paraphrastic agreement across
rewordings to improve answers at inference time. We use paraphrase consistency
to diagnose whether reasoning occurred and thus want it to vary rather than be
maximised; our per-feature decomposition appears in
Appendix~\ref{app:per-feature}.

Related multi-turn agent experiments likewise find that increasing interaction
length improves performance only up to a plateau and can broaden answer coverage
without evidence of deeper inference, illustrating why response length should
not itself be treated as a reasoning signal~\citep{kim_multiturn_2025}.

\citet{bogdan_thought_2025} identify ``thought anchors,'' influential steps
within visible CoT traces. We apply related intervention logic to probe CoT-like
behavior under neutral prompting, using anchor perturbation as one feature rather
than the end goal.

\section{Method}
\label{sec:method}

We operationalize hidden CoT detection as comparative inference. For each model $m$ and question $q$, we evaluate \textit{explicit\_cot}, which requests step-by-step reasoning; \textit{explicit\_no\_cot}, which requests a direct answer; and \textit{neutral\_strict}, a task-only prompt without reasoning-display instructions. For Thinking, neutral and no-CoT additionally share the force-closure intervention below. Appendix~\ref{app:prompts} gives the full prompts. The explicit conditions define CoT and no-CoT \emph{poles}. Rather than recover an internal trace, HCDS asks which pole neutral behavior resembles. Its measured construct is therefore neutral-prompt CoT-like alignment; hidden CoT is the hypothesis it probes. Figure~\ref{fig:methods} in Appendix~\ref{app:hparams} summarizes the pipeline.

\subsection{HCDS}
\label{subsec:hcds}

For each $(m,q,p)$ tuple, we construct a six-dimensional feature vector
\[
\begin{aligned}
f_{m,q,p} = \bigl[&\ell_{m,q,p},\; \bar{H}_{m,q,p},\; s^{H}_{m,q,p},\; C^{\mathrm{para}}_{m,q,p},\\
&\Delta A^{\mathrm{pert}}_{m,q,p},\; \Delta A^{\mathrm{mech}}_{m,q,p}\bigr].
\end{aligned}
\]
where the components denote latency per output token, mean token entropy, entropy slope, paraphrase consistency, perturbation sensitivity, and mechanistic intervention sensitivity. The linguistic and behavioral features characterize prompt-condition differences; the final feature measures interventions on candidate reasoning-related internal states. Primary HCDS retains all six.

Preprocessing is separate by backend and $(\mathrm{model},\mathrm{dataset})$.
Raw features are z-scored for the unadjusted score; for the primary score, raw
features are first length-residualised and those residuals are then z-scored.
We calculate Euclidean distance over the features available in each pair:
\[
D_{\mathcal{J}_q}(a,b) = \sqrt{\sum_{j \in \mathcal{J}_q}(a_j-b_j)^2},
\]
where $\mathcal{J}_q$ is the pair-specific feature intersection, so the two
distances can use different sets. Because the mechanistic feature is sometimes
undefined, the primary \emph{rescale} policy uses
\[
D(a,b) = \sqrt{\frac{|\mathcal{F}|}{|\mathcal{J}_q|}}\,
D_{\mathcal{J}_q}(a,b),
\]
where $\mathcal{F}$ is the full six-feature set.

The \emph{pairwise} policy omits the correction; \emph{complete} uses one
three-pole intersection for both distances. All GSM8K cells remain significant
($p<10^{-3}$); the largest policy difference is $0.40$ HCDS.

We define per-question HCDS as
\begin{equation}
\begin{aligned}
\mathrm{HCDS}_q ={}&
D\!\left(f_{m,q,\mathrm{neutral}},f_{m,q,\mathrm{no\text{-}cot}}\right)\\
&-
D\!\left(f_{m,q,\mathrm{neutral}},f_{m,q,\mathrm{cot}}\right).
\end{aligned}
\label{eq:hcds-q}
\end{equation}

Thus, $\mathrm{HCDS}_q>0$ indicates that neutral behavior is closer to explicit CoT than explicit no-CoT; $\mathrm{HCDS}_q<0$ indicates the reverse. Positive HCDS is evidence consistent with latent intermediate reasoning under the neutral prompt but does not establish causality.

We aggregate per-question scores by model and dataset:
\begin{equation}
\overline{\mathrm{HCDS}} =
\frac{1}{N}\sum_{q=1}^{N}\mathrm{HCDS}_q.
\label{eq:hcds-mean}
\end{equation}

\subsection{Adjusting for Output Length}
\label{subsec:length-adjust}

Several HCDS features use generated tokens, whose counts can differ substantially across prompt conditions. Because sequence length can affect latency and token statistics independently of reasoning, we adjust every feature for generation length before computing primary HCDS.

Within each backend and $(\mathrm{model},\mathrm{dataset})$ group, we fit a
separate OLS regression of each raw feature on $\log n^{\mathrm{gen}}_{m,q,p}$
over all defined question--prompt rows, retain the residual, and z-score those
residuals before applying Equation~\ref{eq:hcds-q}. Output length is only the
adjustment covariate and a diagnostic and matching variable.

We interpret the adjusted score as alignment not linearly explained by
$\log n^{\mathrm{gen}}$, while noting that residualisation may also remove
reasoning-related variation (Section~\ref{sec:limitations}).

\begin{table}[t]
\centering
\small
\caption{\textbf{Degenerate no-CoT pole.} Generated token counts ($p5$ / $p50$ / $p95$) across prompt conditions on StrategyQA ($n=50$). The forced no-CoT condition exhibits zero variance (flat at 5--6 tokens across all percentiles), with zero distributional overlap against CoT and neutral conditions.}
\label{tab:degenerate-pole}
\begin{tabular}{lccc}
\toprule
Backend / Model & \texttt{explicit\_cot} & \texttt{explicit\_no\_cot} & \texttt{neutral\_strict} \\
\midrule
PyTorch/CUDA Instruct & 223 / 372 / 715 & \textbf{6 / 6 / 6} & 137 / 254 / 562 \\
PyTorch/CUDA Thinking & 808 / 1908 / 4466 & \textbf{6 / 6 / 6} & 126 / 1012 / 3615 \\
Apple MLX Instruct    & 218 / 374 / 747 & \textbf{5 / 5 / 5} & 118 / 220 / 562 \\
Apple MLX Thinking    & 898 / 1806 / 4962 & \textbf{5 / 5 / 5} & 115 / 1057 / 3086 \\
\bottomrule
\end{tabular}
\end{table}

\subsection{Establishing a Clean No-CoT Pole}
\label{subsec:clean-pole}

Equation~\ref{eq:hcds-q} requires genuinely different CoT and no-CoT reasoning conditions. This is nontrivial for reasoning-tuned checkpoints whose decoding format automatically opens a \emph{reasoning block}, a special-token-delimited span of intermediate steps before the final answer~\citep{yang_qwen3_2025}.

We use Qwen3-4B Instruct and Thinking, which share an architecture, tokenizer, and pre-training corpus but differ in post-training~\citep{yang_qwen3_2025}. For Thinking, the \texttt{explicit\_no\_cot} instruction does not reliably suppress reasoning: its chat template unconditionally opens a \texttt{\textless think\textgreater} block. We therefore \emph{force-close} the no-CoT reasoning block: the assistant prefix supplies a fabricated completed thought (``Okay, I think I have finished thinking.'') followed by \texttt{\textless/think\textgreater}, so decoding begins with the answer. Following \emph{NoThinking}~\citep{ma2025nothinking}, this produces an answer-only condition without relying on an instruction the checkpoint disobeys.

We apply the same force-closed prefix to \texttt{explicit\_no\_cot} and \texttt{neutral\_strict}. This symmetry controls the modified reasoning format: intervening only on the no-CoT pole would create systematic formatting and verbosity differences between poles. Explicit CoT is generated naturally.

This symmetry is empirically consequential. With an un-prefilled neutral pole,
the unadjusted HCDS fails a basic discriminant-validity check: single-step
arithmetic scores $+2.146$, more than twice GSM8K's $+1.018$. Under symmetric
force-closure, the ordering reverses, with GSM8K at $+2.372$ and arithmetic at
$+1.174$ (Appendix~\ref{app:output-length}). We therefore treat force-closure
as a load-bearing component of the Thinking-model design rather than an
optional implementation detail.

\subsection{Mechanistic Intervention}
\label{subsec:mechanistic}

The mechanistic feature measures whether suppressing candidate reasoning-related internal states changes accuracy relative to matched control interventions. In Qwen3's Transformer architecture, attention and feedforward blocks update a \emph{residual stream}, the running hidden representation passed between layers~\citep{vaswani_attention_2023,elhage_toy_2021}. We identify candidate states through gradient$\times$activation attribution toward the answer span: the elementwise product of an internal activation and the target answer score's gradient with respect to it. Its magnitude estimates first-order output sensitivity. Attribution nominates influential states but is not itself treated as causal evidence.

A \emph{reasoning step} is a contiguous generated-token span treated as one trace unit. We select the highest-attribution steps as candidate reasoning anchors and compare their intervention effects with position-matched controls from the same trace. Appendix~\ref{app:mech-details} describes attribution, segmentation, selection, intervention, and controls.

For each selected step, we perform an \emph{ablation}, replacing the relevant internal activation with zero before continuing generation~\citep{meng_locating_2023}. The mechanistic feature records the accuracy cost relative to the matched control intervention. A larger accuracy decrease after suppressing a candidate reasoning step provides evidence that its internal state causally contributes to the output.

\section{Experimental Setup}
\label{sec:experimental-setup}

We evaluate two \emph{open-weight} Qwen3 models: Qwen3-4B-Instruct-2507 and Qwen3-4B-Thinking-2507~\citep{yang_qwen3_2025}. Instruct is optimized for instruction following; Thinking is post-trained to generate an explicit reasoning block before answering. We test both on GSM8K grade-school word problems requiring multi-step arithmetic~\citep{cobbe_training_2021} and StrategyQA binary questions requiring implicit multi-hop commonsense reasoning~\citep{geva_did_2021}. Both datasets use \textit{explicit\_cot}, \textit{explicit\_no\_cot}, and \textit{neutral\_strict}, hereafter explicit CoT, explicit no-CoT, and neutral.

Within each $(\mathrm{model},\mathrm{dataset})$ cell, the three prompt conditions share identical deterministic decoding settings, including the token cap; caps differ across models and datasets.

For each $(m,q,p)$ tuple, we record answer accuracy, generation time, output length, latency per output token, token-level entropy, paraphrase consistency, perturbation sensitivity, and, when available, mechanistic intervention sensitivity. Accuracy and output length are auxiliary diagnostics excluded from HCDS; the other six measurements form the vector in Section~\ref{sec:method}. Latency per output token is generation time divided by output-token count; mean token entropy averages predictive entropy over the continuation. Entropy slope is a linear trend fitted to token-level entropy over the continuation. Full definitions and preprocessing appear in
Appendices~\ref{app:paraphrase-audit},~\ref{app:hparams}, and~\ref{app:mech-details}.

The content features capture complementary behavioral sensitivities. \emph{Paraphrase consistency} $C^{\mathrm{para}}$ is the fraction of $K$ meaning-preserving paraphrases yielding the original question's answer. Paraphrases preserve all quantities, problem structure, named entities, and the gold answer. \emph{Perturbation sensitivity} is
\[
\Delta A^{\mathrm{pert}}
=
A_{\mathrm{orig}}-A_{\mathrm{pert}},
\]
where the fixed clause ``The neighbor's dog barked seven times that afternoon.''
is prepended to every question; it is neither sampled nor model-generated.
\emph{Mechanistic sensitivity} is
\[
\Delta A^{\mathrm{mech}}
=
A_{\mathrm{control}}-A_{\mathrm{anchor}},
\]
the additional accuracy loss from suppressing attribution-selected anchors rather than position-matched controls. Appendices~\ref{app:paraphrase-audit} and~\ref{app:mech-details} specify paraphrase generation, perturbation construction, anchor and control selection, and interventions.

We report HCDS by model--dataset pair. Because decoding is deterministic, uncertainty reflects variation across questions rather than generations. We test
\[
H_0:\mathbb{E}[\mathrm{HCDS}_q]=0
\qquad\text{vs.}\qquad
H_1:\mathbb{E}[\mathrm{HCDS}_q]\neq0
\]
using question-level scores, bootstrap confidence intervals, and two-sided significance tests. The 1000 bootstrap replicates (seed 17) resample fixed per-question HCDS values and do not refit pooled preprocessing. Positive HCDS indicates greater alignment with explicit CoT; values near zero indicate comparable alignment with both poles. We assign no universal meaning to HCDS magnitude because its scale depends on the standardized feature space.

We test robustness through leave-one-feature-out, single-feature, and length-matched analyses. For open-weight models, we separately report anchor-versus-control effects as mechanistic validation. Appendices~\ref{app:prompts},~\ref{app:paraphrase-audit},~\ref{app:hparams}, and~\ref{app:mech-details} provide additional decoding, paraphrase, perturbation, and implementation details.

\section{Results}
\label{sec:results}

\begin{figure}[!htbp]
  \centering
  \includegraphics[width=\columnwidth]{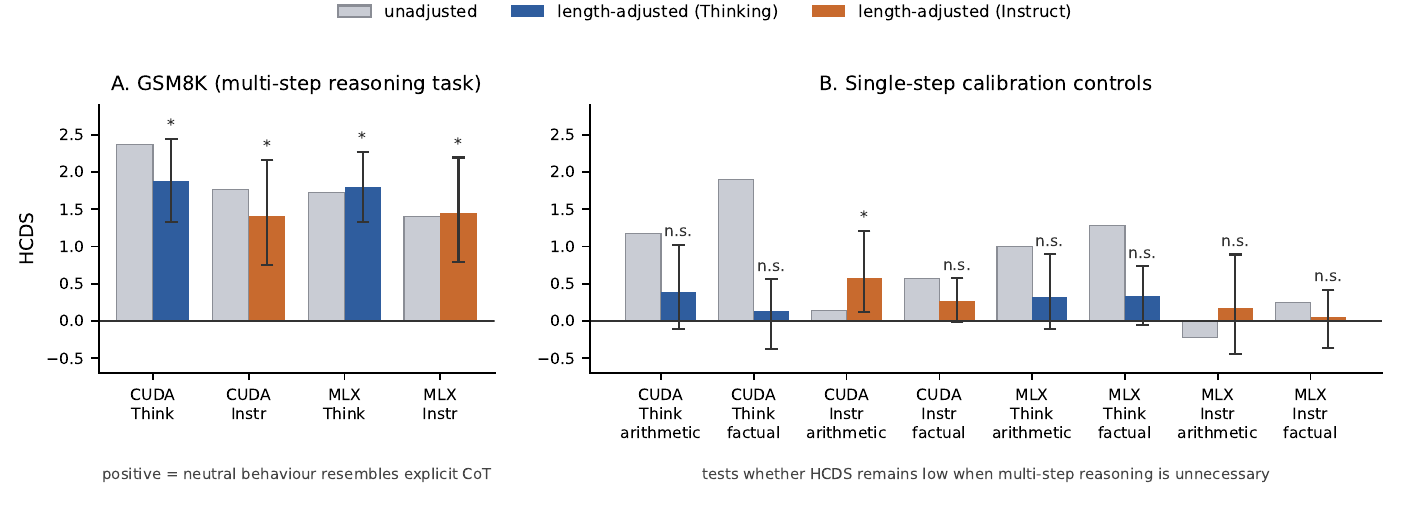}
  \caption{\textbf{HCDS and calibration-control analysis.} $n{=}50$ per cell,
  evaluated independently on PyTorch/CUDA bf16 and MLX 8-bit inference stacks.
  Grey bars show the unadjusted six-feature score; coloured bars show the score
  after residualising each feature on log output length. \textbf{(A)} On GSM8K,
  both scores are positive and significant. \textbf{(B)} On single-step
  arithmetic and factual lookup, the unadjusted score is large in five of eight
  cells, while the adjusted score falls near zero and is non-significant in
  nearly all cells. This motivates the length-adjusted score as our primary
  measure. Error bars and significance markers apply to the coloured adjusted
  bars: intervals are 95\% bootstrap CIs (1000 resamples, seed 17); ${*}$
  denotes $p<0.05$ and n.s. denotes $p\geq0.05$ in a two-sided one-sample
  $t$-test. Unadjusted inference is reported in Table~\ref{tab:negcontrol}.}
  \label{fig:cross-dataset}
\end{figure}

\subsection{HCDS detects CoT-like behavior on GSM8K}

On GSM8K, neutral prompting produces behavior substantially closer to explicit
CoT than explicit no-CoT. After residualising each feature on log output length,
HCDS is positive and highly significant for
both Qwen3-4B variants on both independent inference stacks
(Table~\ref{tab:main-hcds}).

\begin{table}[!h]
\centering
\small
\caption{HCDS on GSM8K, $n{=}50$, before and after length adjustment on two
independent inference stacks. The length-adjusted scores agree across backends
to within $0.08$, compared with a $0.65$ maximum difference before adjustment.}
\label{tab:main-hcds}
\begin{tabular}{llcc}
\toprule
Backend & Model & unadjusted & length-adjusted \\
\midrule
PyTorch/CUDA & Thinking & $+2.372$ & $\mathbf{+1.872}$ \; $[1.32,2.45]$, $p{=}1.2\times10^{-7}$ \\
PyTorch/CUDA & Instruct & $+1.765$ & $\mathbf{+1.410}$ \; $[0.75,2.16]$, $p{=}1.9\times10^{-4}$ \\
MLX 8-bit & Thinking & $+1.722$ & $\mathbf{+1.799}$ \; $[1.33,2.27]$, $p{=}1.3\times10^{-9}$ \\
MLX 8-bit & Instruct & $+1.403$ & $\mathbf{+1.449}$ \; $[0.80,2.19]$, $p{=}1.7\times10^{-4}$ \\
\bottomrule
\end{tabular}
\end{table}

The result replicates across hardware, numerical precision, and runtime, with
substantially tighter agreement after length adjustment
(Table~\ref{tab:main-hcds}), checking runtime-specific artifacts.

\subsection{Length adjustment reduces calibration-control scores}

The unadjusted score cannot reliably distinguish reasoning from verbosity. On
single-step arithmetic and factual lookup, where multi-step reasoning is
unnecessary, it produces large positive values that can exceed those on GSM8K.
After residualising each feature on log output length, control scores approach
zero and are almost uniformly non-significant
(Figure~\ref{fig:cross-dataset}; Table~\ref{tab:negcontrol}). We therefore
interpret the unadjusted score only as a supplementary diagnostic.

\subsection{Mechanistic interventions provide complementary evidence}
\label{sec:results-mech}

The mechanistic feature tests whether attribution-selected intermediate steps
causally affect the final answer by comparing suppression of the
highest-attribution steps with position-matched controls.

The signal is sparse and mixed in sign (33 positive, 37 negative of 70 non-zero CUDA cells)
(Section~\ref{sec:limitations}). The feature therefore contributes little to
the aggregate HCDS: removing it changes the length-adjusted GSM8K score only
modestly (Figures~\ref{fig:ablation} and~\ref{fig:ablation-mlx}).

Full intervention results and selection procedures are provided in
Appendix~\ref{app:mech-details}.

\subsection{The result is robust to feature choice}
\label{sec:robustness}

\begin{figure}[!htbp]
  \centering
  \includegraphics[width=\columnwidth,trim=0 204bp 0 0,clip]{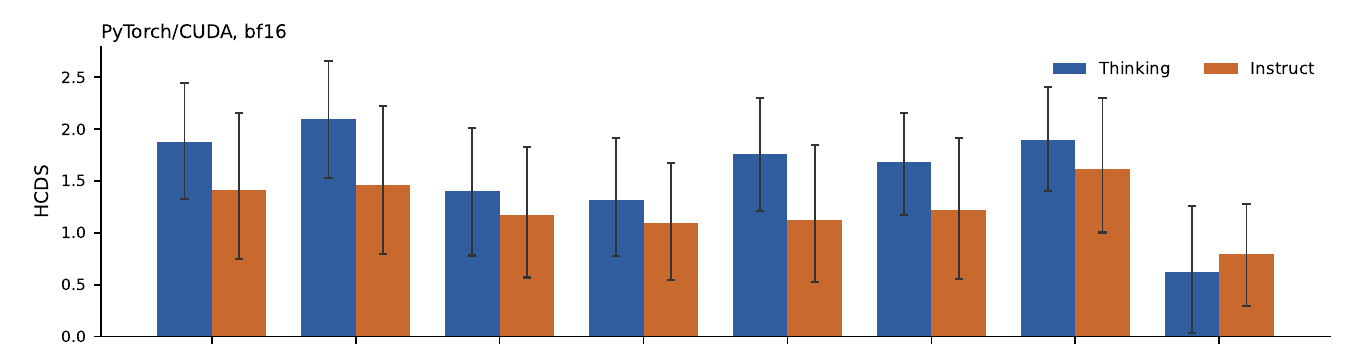}
  \vspace{-0.6ex}

  \makebox[\columnwidth][r]{%
    \includegraphics[width=0.8765\columnwidth,trim=80bp 18bp 0 328.4bp,clip]{figures/fig2_robustness_ablation.pdf}}
  \caption{\textbf{Feature-ablation robustness of the length-adjusted HCDS}
  on GSM8K, $n{=}50$, for PyTorch/CUDA bf16 across both models. Across both
  inference stacks, all 24 leave-one-feature-out variants remain positive and
  significant; the MLX 8-bit arm appears in Figure~\ref{fig:ablation-mlx} in
  Appendix~\ref{app:sensitivity}. Removing both entropy features produces the
  largest reduction, but the resulting score remains significant. Across all
  four model--backend cells, removing latency increases HCDS; only the two CUDA
  cells are displayed here. Error bars show 95\% bootstrap CIs.}
  \label{fig:ablation}
\end{figure}

No single feature drives the GSM8K result. All 24 leave-one-feature-out
variants across two models and two inference stacks remain positive and
significant, with the weakest cell at $+1.193$ ($p=1.6\times10^{-3}$)
(Figures~\ref{fig:ablation} and~\ref{fig:ablation-mlx}). Removing both entropy
features produces the largest reduction, but the score remains significant for
both models on both backends. Thus, although entropy is the strongest
contributor, the headline result does not depend on it.

\paragraph{Per-feature contributions.}

We evaluate which individual features align neutral behavior with explicit CoT
using the one-dimensional contrast
\[
\delta_j(q)
=
\left|z_j^{\mathrm{neu}}-z_j^{\mathrm{no\text{-}cot}}\right|
-
\left|z_j^{\mathrm{neu}}-z_j^{\mathrm{cot}}\right|.
\]
Positive values indicate that neutral behavior is closer to the CoT pole on
feature $j$ alone. Because the full score uses Euclidean distance, these
one-dimensional contrasts do not sum to HCDS.

\begin{table}[!htbp]
\centering
\small
\setlength{\tabcolsep}{5pt}
\begin{tabular}{@{}lrrrrrr@{}}
\toprule
& \multicolumn{3}{c}{Thinking} & \multicolumn{3}{c}{Instruct} \\
\cmidrule(lr){2-4}\cmidrule(lr){5-7}
Feature & $\bar{\delta}_j$ & $d$ & $p$ & $\bar{\delta}_j$ & $d$ & $p$ \\
\midrule
latency & $-0.11$ & $-0.19$ & $0.175$ & $+0.07$ & $+0.49$ & $0.001$ \\
entropy mean & $+1.05$ & $+0.93$ & $2.7\times10^{-8}$ & $+0.70$ & $+0.57$ & $1.8\times10^{-4}$ \\
entropy slope & $+1.22$ & $+1.13$ & $2.0\times10^{-10}$ & $+0.83$ & $+0.56$ & $2.5\times10^{-4}$ \\
paraphrase consistency & $+0.67$ & $+0.66$ & $2.5\times10^{-5}$ & $+0.87$ & $+0.77$ & $1.8\times10^{-6}$ \\
perturbation sensitivity & $+0.33$ & $+0.24$ & $0.101$ & $+0.33$ & $+0.23$ & $0.114$ \\
mechanistic sensitivity & \multicolumn{6}{c}{not estimable} \\
\bottomrule
\end{tabular}
\caption{Per-feature contrast on length-adjusted GSM8K, PyTorch/CUDA,
$n{=}50$. $\bar{\delta}_j$ is the mean one-dimensional contrast in z-scored
units, with positive values indicating greater alignment with explicit CoT.
Mechanistic sensitivity is not estimable because it is available at all three
prompt poles for too few questions.}
\label{tab:per-feature}
\end{table}

Entropy is the strongest and most consistent contributor across both models
($d=0.56$--$1.13$), while paraphrase consistency is the strongest non-entropy
feature, particularly for Instruct. Latency contributes little after
residualisation, consistent with the increase in HCDS observed when it is
removed. Perturbation sensitivity is positive but does not reach significance
in either model on this sample.

\paragraph{Entropy is not explained by length alone.}

Because entropy contributes most strongly, we test whether it duplicates output
length or other behavioral features. VIF values remain below
$1.9$ for all reported predictors, indicating low multicollinearity. After
controlling simultaneously for generation length, decode latency, paraphrase
consistency, and perturbation sensitivity, entropy retains a significant
partial association with the reasoning condition on GSM8K (Thinking,
$p=0.028$). On GSM8K, neutral entropy is also closer to the explicit-CoT pole
than to the no-CoT pole, whereas entropy approaches the no-CoT pole on the
single-step controls.

Thus, entropy contributes information beyond output length and other behavioral
features, but this does not establish that entropy directly measures internal reasoning.

\subsection{Cross-task results}

StrategyQA does not provide an independent replication: after length
adjustment, Instruct is nonsignificant on both backends, while Thinking remains
positive but is driven mainly by entropy slope. Its binary answer format
compresses the accuracy-based features near the chance floor; full results
appear in Appendix~\ref{app:strategyqa}.

\paragraph{Sensitivity to feature weighting.}

Across all $2^6-1=63$ non-empty feature subsets, the GSM8K score remains
positive in $95.2$--$98.4\%$ of subsets and significant in $84.1$--$90.5\%$
across models and backends. Negative subsets are concentrated in degenerate
combinations containing only latency and/or the sparse mechanistic feature.
A correlation-aware Mahalanobis metric leaves the GSM8K conclusions unchanged
(Appendix~\ref{app:mahalanobis}).

\section{Discussion}
\label{sec:discussion}

A key control-task check asks whether length adjustment reduces positive scores on
single-step arithmetic and factual lookup, where multi-step reasoning is unnecessary
(Figure~\ref{fig:cross-dataset}B).

Given a valid no-CoT pole by force-closure, Thinking scores higher than the more
compliant Instruct on both backends (Table~\ref{tab:main-hcds}). Prompt compliance therefore
makes a model easier to measure, not less likely to reason latently.

\begin{table}[!h]
\centering
\small
\caption{Negative-control analysis, full six-feature HCDS, $n{=}50$ per cell.
Bold marks significantly positive cells under the control-task calibration criterion.
The unadjusted score is significantly positive in five of
eight cells; the length-adjusted score is significantly positive in one, marginally.}
\label{tab:negcontrol}
\begin{tabular}{lllcc}
\toprule
Backend & Model & Control task & unadjusted & length-adjusted \\
\midrule
CUDA & Thinking & arithmetic & $\mathbf{+1.174}$ ($p{=}1.7\mathrm{e}{-4}$) & $+0.392$ (n.s.) \\
CUDA & Thinking & factual    & $\mathbf{+1.903}$ ($p{=}2.4\mathrm{e}{-12}$) & $+0.131$ (n.s.) \\
CUDA & Instruct & arithmetic & $+0.143$ (n.s.) & $\mathbf{+0.578}$ ($p{=}0.046$) \\
CUDA & Instruct & factual    & $\mathbf{+0.567}$ ($p{=}3.7\mathrm{e}{-3}$) & $+0.267$ (n.s.) \\
MLX  & Thinking & arithmetic & $\mathbf{+0.996}$ ($p{=}3.0\mathrm{e}{-4}$) & $+0.323$ (n.s.) \\
MLX  & Thinking & factual    & $\mathbf{+1.277}$ ($p{=}5.9\mathrm{e}{-8}$) & $+0.334$ (n.s.) \\
MLX  & Instruct & arithmetic & $-0.223$ (n.s.) & $+0.173$ (n.s.) \\
MLX  & Instruct & factual    & $+0.248$ (n.s.) & $+0.050$ (n.s.) \\
\bottomrule
\end{tabular}
\end{table}

Backend replication cannot remove a length dependence shared by both stacks,
while residualisation addresses it directly; the one marginal exception in
Table~\ref{tab:negcontrol} neither replicates nor survives multiple-comparisons
correction (Appendix~\ref{app:multiplicity}).

\section{Limitations}
\label{sec:limitations}

\paragraph{The Thinking no-CoT pole is obtained under intervention.}
Both Thinking poles in Equation~\ref{eq:hcds-q} use force-closure prefilling
(Section~\ref{subsec:clean-pole}) because this checkpoint has no promptable
non-reasoning mode. Compliance is therefore \emph{engineered rather than
observed}, and the Thinking results characterize behavior under this
intervention. Appendix~\ref{app:output-length} reports the un-intervened
alternative and its aggregate-score failure. Evidence that the prefill does not itself produce the effect comes
from its identical application to both conditions: \texttt{explicit\_no\_cot}
yields a median of 6 tokens versus 1012 for \texttt{neutral\_strict} on
StrategyQA. Thus, the instruction rather than the shared prefix alone explains
the observed output-length difference under this intervention.

\paragraph{Single model family, with a cross-family check.}
Our qualitative comparison of reasoning-tuned and instruction-tuned models uses
one family, Qwen3-4B Instruct and Thinking. Their shared architecture, tokenizer,
and pre-training corpus and differing post-training make the within-family
comparison clean, but prevent us from separating Thinking-class behavior from
this specific Qwen3-4B post-training recipe. As an exploratory cross-family
check, an unadjusted pilot on Gemma-3-4B-it, with different architecture,
tokenizer, and pre-training corpus, yields HCDS $+1.54$ (95\% CI $[1.02,2.06]$,
$n=50$, $p<10^{-6}$) on GSM8K, directionally consistent with the unadjusted
Qwen3-4B Instruct result (Table~\ref{tab:main-hcds}). Because per-question generation lengths were not
recorded, the pilot cannot be length-adjusted, and Gemma-3-4B has no
reasoning-tuned sibling. A key open direction is therefore a full
length-adjusted, dual-variant replication on a second reasoning-tuned family
with a correctly matched base checkpoint (e.g., DeepSeek-R1-Distill paired with
Qwen2.5-Math).

\paragraph{Mechanistic component is preliminary.}
The mechanistic component is the weakest of the six features.
$\Delta A^{\mathrm{mech}}$ ranks steps by
gradient$\times$activation attribution toward the answer span, selects the top
2 per trace as anchors, and compares their suppression with 1 non-anchor step
matched on relative trace position (Appendix~\ref{app:mech-details}). Five
limitations constrain this measure: it uses one pinned
$(\text{layer},\text{site})$ pair rather than a sweep; the answer-containing
step is eligible for selection; controls are position- but not
\emph{function}-matched; gradient$\times$activation is susceptible to
first-order saturation; and coverage is sparse. Specifically,
$\Delta A^{\mathrm{mech}}$ is defined in 488/1200 PyTorch/CUDA cells (41\%) and
482/1200 MLX cells (40\%); among defined cells, 86\% and 94\%, respectively,
are exactly zero, leaving non-zero values in only roughly $2$--$6\%$ of all
cells. This undefinedness is structural rather than missing-at-random: the
contrast requires at least three segmentable reasoning steps and a parseable
answer span, which the terse \texttt{explicit\_no\_cot} condition often lacks.
The feature therefore contributes little to HCDS, as confirmed by ablation, but
we retain it for construct validity because HCDS is intended to combine
behavioral and internal evidence.
A standalone anchor-suppression study with a full layer/site sweep,
function-matched controls, and a less saturation-prone attribution method is
therefore the priority mechanistic follow-up. We make no claim about \emph{which}
reasoning steps carry causal load.

\paragraph{Statistical reporting.}
Deterministic decoding with one trial per question makes question identity,
rather than model stochasticity, the unit of statistical variation. The
reported uncertainty therefore reflects between-question variability in HCDS,
not sampling uncertainty about underlying model behavior. Because the bootstrap
resamples fixed per-question scores rather than refitting preprocessing, it also
conditions on the fitted residualisation and standardisation. The six features in
$f_{m,q,p}$ are also derived from the same response and may be autocorrelated
through output length. We confirmed substantial length dependence, motivating
residualisation of every feature on $\log$ generation length
(Section~\ref{subsec:length-adjust}). A length-matched analysis is reported in
Appendix~\ref{app:length-matched}. However, residualisation cannot distinguish
confounding from mediation: if
neutral prompting causes hidden reasoning that causes longer outputs, adjustment
removes genuine signal along with the artifact. The adjusted scores in
Table~\ref{tab:main-hcds} measure alignment not linearly explained by $\log$
generation length; they do not separate confounding from mediation.
Distinguishing these cases requires intervening directly on length rather than
adjusting for it. Stochastic multi-trial evaluation would additionally quantify
within-prompt variance.

\section{Conclusion}

HCDS is a comparative assay of neutral-prompt CoT-like alignment combining linguistic, behavioral, and mechanistic indicators. It asks whether neutral behavior is closer to overt CoT or no-CoT in a six-feature space; it neither directly observes an internal trace nor establishes the causal role of visible reasoning. On GSM8K, length-adjusted HCDS is robustly positive for both Qwen3-4B variants across hardware and precision stacks (Table~\ref{tab:main-hcds}) and substantially reduces positive scores on calibration controls (Table~\ref{tab:negcontrol}). Instruct follows answer-only directives naturally; Thinking requires force-closure for a clean baseline. Under that shared intervention, Thinking complies when instructed but produces lengthy CoT-like behavior when neutral, a pattern consistent with less prompt-conditional reasoning but not proof of an unexposed trace.

Key open questions are whether this distinction generalizes beyond Qwen3-4B and whether a full layer/site sweep with function-matched controls can establish \emph{which} reasoning steps carry causal load, which our mechanistic feature is too coarse to answer.

\bibliography{references}
\bibliographystyle{plainnat}

\clearpage
\appendix

\section{Prompt Templates}
\label{app:prompts}

All three conditions use the same system-plus-user chat-template structure, but
not the same answer-format instruction: the two explicit conditions request a
boxed answer, whereas the intentionally minimal neutral condition does not.
The blocks below show GSM8K. StrategyQA uses the identical text with
``numeric/number'' replaced by ``yes or no/yes-no'' and
\texttt{\textbackslash boxed\{number\}} replaced by
\texttt{\textbackslash boxed\{yes/no\}}. System messages are otherwise fixed;
only the user message substitutes the question.

\promptblock{explicit\_cot}
  {Think through this step by step, show your reasoning process, then
  provide your final answer. Put the final numeric answer in
  \texttt{\textbackslash boxed\{number\}}.}
  {Question: $\langle q\rangle$}

\promptblock{explicit\_no\_cot}
  {Answer-only mode. \texttt{/no\_think}\newline
  Output exactly one line: \texttt{\textbackslash boxed\{number\}}. Do not include
  reasoning, explanations, equations, restatements, or units.}
  {Question: $\langle q\rangle$\newline
  \texttt{/no\_think}\newline
  Respond only with \texttt{\textbackslash boxed\{number\}}.}

\promptblock{neutral\_strict}
  {You are a helpful assistant.}
  {Question: $\langle q\rangle$}

The intentionally minimal \texttt{neutral\_strict} prompt contains no reasoning
or answer-format directive. Answer extraction therefore falls back to
``last number in text'' when no \texttt{\textbackslash boxed\{\}} delimiter is produced.

Paraphrases for the paraphrase-consistency feature are generated by a separate
call to the Instruct checkpoint, with the question text as the sole user turn:

\promptblock{paraphrase generation}
  {Rewrite the following question in different words while preserving every
  number, the underlying structure, any named entities, and the correct answer
  exactly. Output only the rewritten question, nothing else.}
  {$\langle q\rangle$}

This system message instructs the generator to preserve, but does not state, the
correct answer; the gold value never enters its context. Unlike the three
evaluation conditions, paraphrase generation samples rather than decoding
greedily ($T = 0.8$, top-$p = 0.9$, seed $17 + k$ for the $k$-th of $K{=}2$
draws, capped at 96 new tokens). Appendix~\ref{app:paraphrase-audit} audits
what those settings produce.

\section{Paraphrase Generation and Quality Audit}
\label{app:paraphrase-audit}

Paraphrase consistency is one of HCDS's two largest contributors
(Appendix~\ref{app:per-feature}), warranting an audit of all $400$ generated
paraphrases across the four final evaluation sets. We exclude the superseded
$n=20$ GSM8K file, which duplicates the first 20 entries of the final $n=50$
file.

\paragraph{The generator never sees the gold answer.} Paraphrases are produced
by the Instruct checkpoint from one system instruction --- reproduced in
Appendix~\ref{app:prompts} --- that asks it to preserve every number, the
underlying structure, named entities, and the correct answer, with the question
text as the sole user turn. The instruction refers to but never states the
correct answer, so the gold label never enters the generator's context. This
prevents direct gold-label leakage, but not \emph{reasoning leakage}: a
paraphrase can simplify a relation (for example, replacing ``five pen pals and
stopped writing to two'' with ``the remaining 3''). Paraphrase consistency is
therefore a noisy behavioral feature rather than a leakage-free measure.

\paragraph{Semantic drift is rare.} After normalising thousands separators and
trailing punctuation, we compare each paraphrase's multiset of numeric values
with its original. Thirteen of $400$ ($3.25\%$) differ. They encode equivalent
quantities, although some simplify relations: \emph{a decade} rewritten as
\emph{10 years}, \emph{triple digits} as \emph{above 99}, \emph{quarters} as
\emph{15-minute quarters}.

\paragraph{Two degenerate cases matter more than drift.} Six of $400$ paraphrases
($1.5\%$) end mid-sentence, all on GSM8K, whose questions are the longest. More consequentially, sampling at
$T=0.8$ with $K=2$ frequently returns the same rewording twice: this happens
for $138$ of the $200$ questions ($69\%$, and $96\%$ on the arithmetic control),
which reduces the effective $K$ to one and makes the feature ternary in
practice. The observed values are exactly $\{0, 0.5, 1\}$; $248$ of $300$ CUDA
and $245$ of $300$ MLX GSM8K cells take the value $1$.

Eight GSM8K questions have byte-identical rewordings. Under greedy decoding,
these comparisons are consistent by construction: the feature is $1.000$ in
all $24$ affected cells per model and backend. Elsewhere, its means are $0.790$
and $0.849$ for CUDA Thinking and Instruct, and $0.798$ and $0.833$ for MLX.

\paragraph{Which way the artifact pushes.} Because those eight questions score
$1.000$ under \emph{every} prompt, they contribute nothing to the
per-question contrast on that feature: they dilute the paraphrase dimension
rather than biasing it. Recomputing length-adjusted GSM8K without them confirms
the direction: HCDS rises in all four cells, by $+0.009$
(MLX Thinking), $+0.043$ (CUDA Thinking), $+0.044$ (MLX Instruct) and $+0.128$
(CUDA Instruct). Thus, in this diagnostic, removing the affected questions
increases rather than decreases the reported values. We retain the full
$n{=}50$ everywhere rather than
dropping questions on a post-hoc quality criterion, and note the effect here.

\section{Decoding and Hyperparameters}
\label{app:hparams}

\begin{table}[!htbp]
\caption{Decoding and statistical settings. Primary $n=50$ baseline token caps
vary by backend, model, and dataset. The separate un-intervened GSM8K $n=500$
output-length diagnostic used a 1024-token cap. Decoding mode, random seed, and
trial count are uniform, and the question is the unit of statistical variation.}
\label{tab:hparams}
\centering
\small
\setlength{\tabcolsep}{4pt}
\renewcommand{\arraystretch}{1.08}
\begin{tabularx}{0.82\linewidth}{@{}>{\raggedright\arraybackslash}p{0.48\linewidth}Y@{}}
\toprule
Setting & Value \\
\midrule
Decoding & Greedy ($\mathrm{do\_sample}{=}\mathrm{False}$) \\
Random seed & 17 \\
Trials per cell & 1 \\
Backends & PyTorch/CUDA bf16; MLX 8-bit (\texttt{Qwen3-4B-*-MLX-8bit}) \\
Primary $n=50$ cap, CUDA StrategyQA & Thinking 6144; Instruct 2048 \\
Primary $n=50$ cap, CUDA GSM8K/arithmetic/factual & Thinking 3072; Instruct 1024 \\
Primary $n=50$ cap, MLX all datasets & Thinking 6144; Instruct 2048 \\
$n=500$ output-length diagnostic & PyTorch/CUDA; natural/un-prefilled; 1024 \\
Max new tokens (mech baseline) & 1024 \\
Max new tokens (post-intervention redecoding) & 1024 \\
Bootstrap resamples & 1000 \\
Bootstrap seed & 17 \\
$t$-test & one-sample, two-sided \\
\bottomrule
\end{tabularx}
\end{table}

\begin{figure}[!htbp]
\refstepcounter{pipelinefigure}
\label{fig:methods}
\centering
\includegraphics[width=\linewidth]{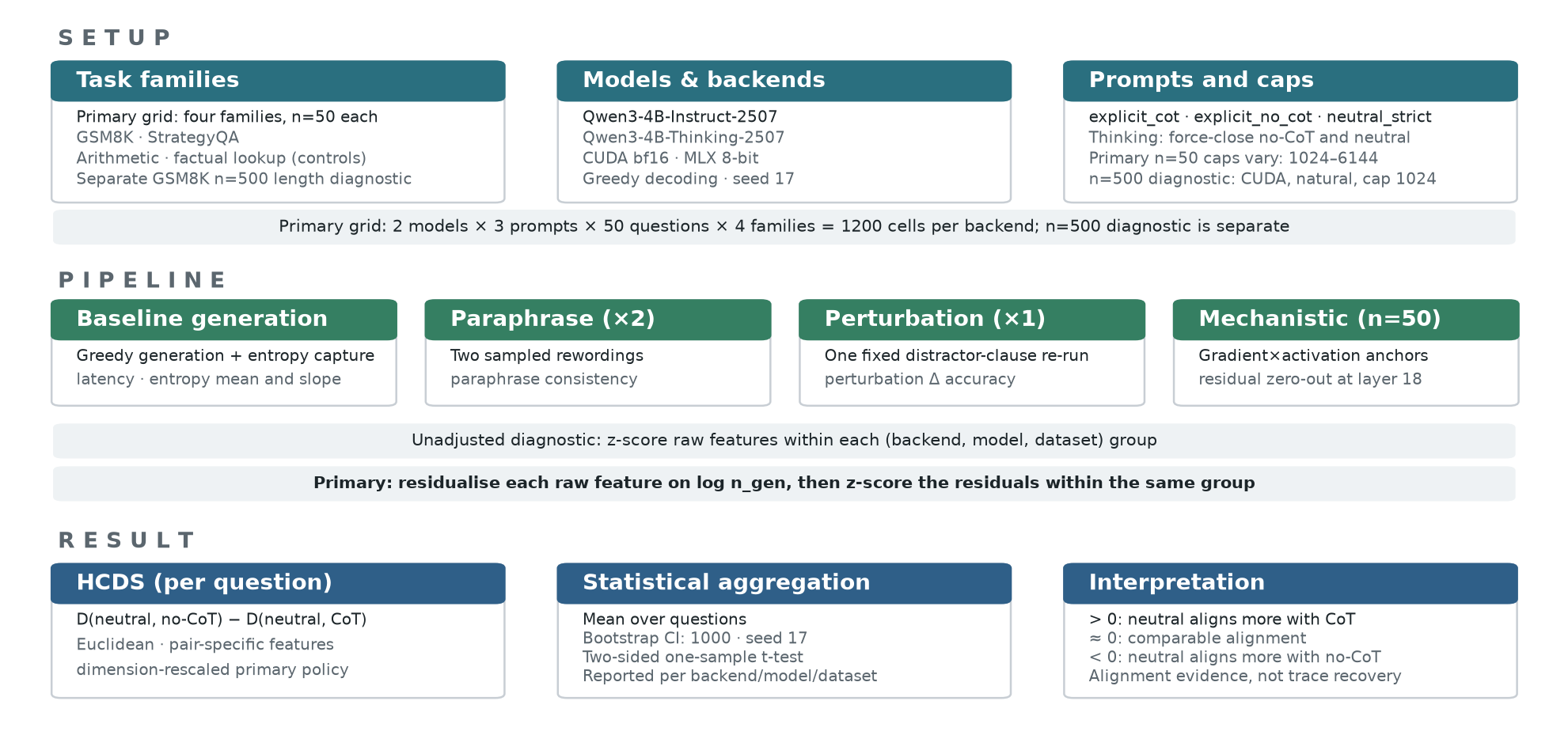}
\caption*{\textbf{Figure~\thepipelinefigure: HCDS methodology pipeline.}
\textit{Setup:} a primary $n=50$ grid of four task families $\times$ two models
$\times$ three prompts, plus a separate $n=500$ output-length diagnostic.
\textit{Pipeline:} for each primary $(\mathrm{model},\mathrm{prompt},\mathrm{question})$ cell,
four parallel sub-pipelines produce a six-dimensional feature vector. Raw
features (unadjusted analysis) or length residuals (primary analysis) are
z-scored within each backend, model, and dataset.
\textit{Result:} HCDS contrasts the neutral prompt's distance to the no-CoT pole
with its distance to the CoT pole. Positive HCDS indicates greater alignment with
explicit CoT, while negative HCDS indicates greater alignment with explicit no-CoT.
Statistical inference is described in Section~\ref{sec:experimental-setup}.}
\end{figure}

\section{Output-Length Distributions}
\label{app:output-length}

\begin{figure}[!htbp]
  \centering
  \includegraphics[width=\columnwidth]{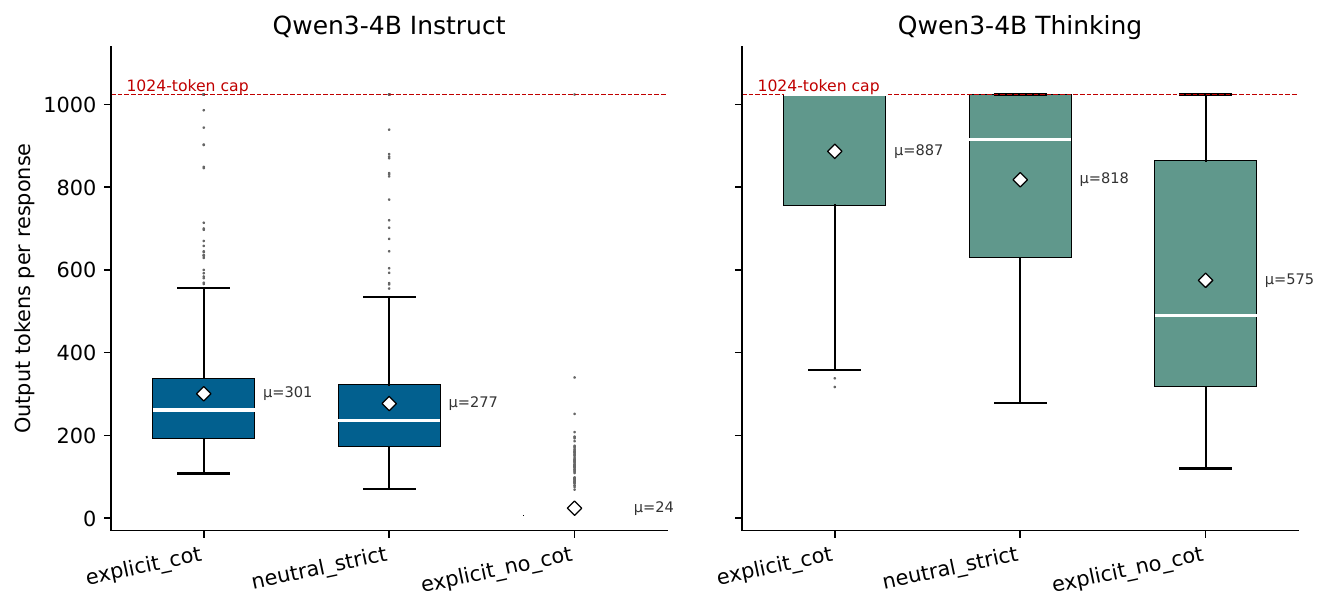}
  \caption{\textbf{Output-length distributions in the separate PyTorch/CUDA, un-intervened GSM8K $n{=}500$ diagnostic.}
  This diagnostic used natural/un-prefilled prompt conditions and a 1024-token
  cap; it is not the primary force-closed $n=50$ HCDS experiment.
  Qwen3-4B-Instruct approaches an answer-only regime under
  \texttt{explicit\_no\_cot} (mean $\mu=24$ tokens), whereas
  Qwen3-4B-Thinking still produces long continuations in the same condition
  (mean $\mu=575$; median 490.5). The \texttt{neutral\_strict}
  distribution remains close to explicit CoT for Thinking and closer to CoT
  than to no-CoT for Instruct. The plot is descriptive and is not used to
  compute the primary HCDS values.}
  \label{fig:output-length}
\end{figure}

\begin{table}[!htbp]
\centering
\small
\caption{Unadjusted six-feature HCDS for Thinking under the two neutral-pole
designs. With the natural (un-prefilled) neutral condition the score is inverted
--- trivial arithmetic outranks GSM8K --- so that design does not measure
task-dependent reasoning. $n{=}50$ per cell.}
\label{tab:neutral-design}
\begin{tabular}{lcccc}
\toprule
Neutral pole & GSM8K & StrategyQA & Arithmetic & Factual \\
\midrule
natural (un-prefilled) & $+1.018$ & $+1.486$ & $\mathbf{+2.146}$ & $\mathbf{+2.132}$ \\
force-closed           & $+2.372$ & $+2.567$ & $+1.174$ & $+1.903$ \\
\bottomrule
\end{tabular}
\end{table}

Table~\ref{tab:neutral-design} confirms this aggregate-score failure.

In the separate diagnostic shown in Figure~\ref{fig:output-length}, the natural
\texttt{explicit\_no\_cot} instruction sharply compresses Instruct outputs but
does not suppress Thinking's long-form generation. The primary analysis instead
uses the symmetric force-closure design in Table~\ref{tab:neutral-design}.

\section{Negative-Control Calibration}
\label{app:negative-control}

A natural concern is that HCDS detects prompt-style differences---longer,
higher-entropy neutral than no-CoT completions---rather than reasoning. We
therefore ran HCDS on two task families where
multi-step hidden reasoning is implausible:

\begin{itemize}
\item \textbf{Single-step arithmetic} ($n=50$): randomly generated
  questions like ``What is $8 \times 6$?''. Gold answers are
  programmatic.
\item \textbf{Numeric-answer factual lookup} ($n=50$): hand-curated
  well-known facts with verifiable integer answers (``How many
  states are in the US?'', ``In what year did humans first land on
  the Moon?'').
\end{itemize}

Both families use the same baseline pipeline, three prompts, and two variants
(Appendix~\ref{app:hparams}). Table~\ref{tab:negcontrol} gives their primary,
six-feature negative-control analysis.

\section{StrategyQA as a Secondary Benchmark}
\label{app:strategyqa}

We report StrategyQA separately from GSM8K rather than as a co-primary
result. The reason is structural: StrategyQA has boolean answers, so
chance accuracy is $0.50$, and the accuracy-difference features in
$f_{m,q,p}$ have correspondingly little dynamic range on it.

\begin{table}[!h]
\centering
\small
\caption{Accuracy by prompt condition and mechanistic coverage. On
StrategyQA the \texttt{explicit\_no\_cot} pole sits close to the $0.50$
chance floor, compressing the usable range of every accuracy-based
feature; on GSM8K, where chance is effectively zero, the same pole is
far from the floor.}
\label{tab:sqa-floor}
\begin{tabular}{llcccc}
\toprule
Model & Dataset & \texttt{cot} & \texttt{no\_cot} & \texttt{neutral} & mech.\ defined \\
\midrule
Instruct & GSM8K      & 0.40 & 0.20 & 0.38 & 81/150 \\
Instruct & StrategyQA & 0.76 & 0.66 & 0.56 & 50/150 \\
Thinking & GSM8K      & 0.40 & 0.14 & 0.38 & 89/150 \\
Thinking & StrategyQA & 0.78 & 0.56 & 0.62 & 50/150 \\
\bottomrule
\end{tabular}
\end{table}

Two consequences follow. First, mechanistic coverage on StrategyQA is
$50/150$ cells against $81$--$89/150$ on GSM8K, because the
anchor-suppression contrast is undefined for traces with fewer than
three steps or no parseable answer span. Second, and more importantly,
once output length is adjusted for (Section~\ref{sec:method}) the
StrategyQA signal is carried almost entirely by a single feature,
entropy slope: in a leave-one-feature-out analysis, removing entropy
slope costs $-0.60$ HCDS while every other feature contributes
$|\Delta| < 0.21$. On GSM8K the adjusted signal remains genuinely
multi-feature, with entropy mean, entropy slope, paraphrase consistency
and perturbation sensitivity all contributing.

\begin{table}[!h]
\centering
\small
\caption{StrategyQA HCDS, $n{=}50$, before and after length adjustment,
on both backends. The Instruct result does not survive length
adjustment on either backend.}
\label{tab:sqa-hcds}
\begin{tabular}{llcc}
\toprule
Backend & Model & unadjusted & length-adjusted \\
\midrule
PyTorch/CUDA & Thinking & $+2.567$ ($p{=}9.5\mathrm{e}{-18}$) & $+0.724$ ($p{=}3.8\mathrm{e}{-4}$) \\
PyTorch/CUDA & Instruct & $+2.174$ ($p{=}9.2\mathrm{e}{-13}$) & $+0.291$ (n.s., $p{=}0.28$) \\
MLX 8-bit    & Thinking & $+1.946$ ($p{=}2.0\mathrm{e}{-12}$) & $+0.738$ ($p{=}1.2\mathrm{e}{-3}$) \\
MLX 8-bit    & Instruct & $+0.863$ ($p{=}2.1\mathrm{e}{-4}$)  & $-0.149$ (n.s., $p{=}0.54$) \\
\bottomrule
\end{tabular}
\end{table}

The negative result is plain (Table~\ref{tab:sqa-hcds}):
\textbf{Instruct StrategyQA HCDS does not survive length adjustment}, on
either backend. This does not show that Instruct performs no hidden reasoning on
commonsense questions; instead, StrategyQA is a weak instrument for the reasons
in Table~\ref{tab:sqa-floor}. The all-subsets sweep agrees: on the
adjusted score, Instruct is significantly positive in $0$ of $63$ feature
subsets on both backends, and Thinking in only 52.4\% (CUDA) and 57.1\%
(MLX), against 84.1\%--90.5\% for every GSM8K cell. Thinking survives on both
backends with close agreement ($+0.724$ vs $+0.738$) but rests on one feature
and should be weighted accordingly. A benchmark with open-ended answers and a low chance floor
would be a better second dataset, and we mark that as future work.

\section{Length-Matched HCDS}
\label{app:length-matched}

\begin{figure}[!t]
\centering
\includegraphics[width=0.92\columnwidth]{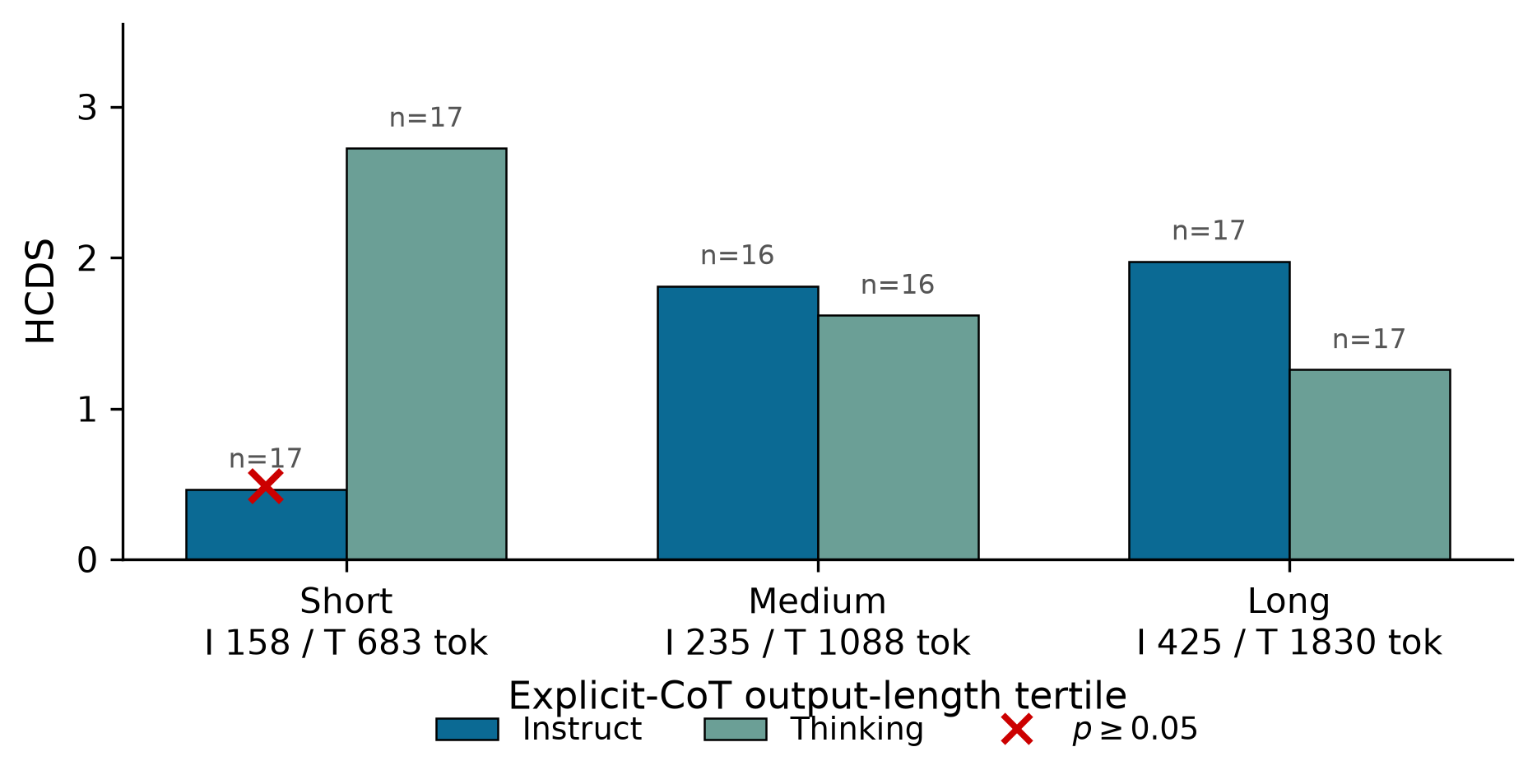}
\caption{\textbf{Length-matched HCDS by output-length tier on GSM8K, $n{=}50$, PyTorch/CUDA, length-adjusted.}
Both models are positive across all tiers, with Instruct short the only
non-significant cell (red $\times$). Tick labels give current mean
\textit{explicit\_cot} token counts for Instruct (I) and Thinking (T).}
\label{fig:length}
\end{figure}

We split GSM8K $n{=}50$ into per-model tertiles of \textit{explicit\_cot}
output length and recomputed HCDS in each;
Figure~\ref{fig:length} and Table~\ref{tab:length-matched} report the results.
HCDS is positive in all six tiers and significant in five; the exception is
Instruct short ($+0.465$, $p=0.23$), the tier with the shortest outputs and
correspondingly least room for the contrast. Thinking is positive and
significant across all three tiers, with lower scores in the medium and long
tiers than in the short tier; the cause of this pattern is not established.
Unadjusted values are uniformly stronger,
ranging from Instruct short $+0.781$ ($p=5.4\mathrm{e}{-2}$) to Thinking short
$+3.286$ ($p=8.6\mathrm{e}{-7}$).

\begin{table}[!htbp]
\caption{Length-matched HCDS by output-length tier on GSM8K, $n{=}50$,
PyTorch/CUDA, length-adjusted. Tier boundaries are per-model tertiles of
explicit-CoT output length. Unadjusted values are uniformly higher and are
given in the text.}
\label{tab:length-matched}
\centering
\small
\setlength{\tabcolsep}{3pt}
\renewcommand{\arraystretch}{1.08}
\begin{tabular}{@{}llrrrr@{}}
\toprule
Model & Tier & $n$ & CoT tok & HCDS & $p$ \\
\midrule
Instruct & short  & 17 & 157.9  & $+0.465$ & $2.3\mathrm{e}{-1}$ (n.s.) \\
Instruct & medium & 16 & 234.9  & $+1.813$ & $2.0\mathrm{e}{-2}$ \\
Instruct & long   & 17 & 425.0  & $+1.976$ & $9.7\mathrm{e}{-3}$ \\
Thinking & short  & 17 & 682.9  & $+2.726$ & $7.2\mathrm{e}{-6}$ \\
Thinking & medium & 16 & 1088.3 & $+1.618$ & $2.4\mathrm{e}{-2}$ \\
Thinking & long   & 17 & 1830.1 & $+1.257$ & $1.4\mathrm{e}{-2}$ \\
\bottomrule
\end{tabular}
\end{table}

\section{Mechanistic Anchor Methodology --- Full Details}
\label{app:mech-details}

This appendix provides the details needed to evaluate or reimplement the
mechanistic feature $\Delta A^{\mathrm{mech}}$ summarized in
Section~\ref{sec:results-mech}.

\paragraph{Step segmentation and the answer span.}
A generated continuation is split into trace steps at sentence and line
boundaries, cutting at each \texttt{[.!?]\textbackslash s+}
or \texttt{\textbackslash n+}. We discard empty chunks and map each
surviving chunk back to a contiguous token-index range using the
tokenizer's offset mapping. The \emph{answer span} is the token range of
the \emph{last} \texttt{\textbackslash boxed\{\ldots\}} match in the
continuation. The step containing that span is not excluded from anchor
selection, so an anchor can be answer-proximal rather than an intermediate
reasoning step. A cell enters the anchor analysis only if it has a
parseable answer span \emph{and} at least three segmented steps. This gate makes
the feature undefined on much of the grid: the terse
\texttt{explicit\_no\_cot} pole
produces neither multiple steps nor, in the Instruct case, anything
beyond a bare boxed answer, so $\Delta A^{\mathrm{mech}}$ is structurally
undefined there rather than missing at random.

\paragraph{Anchor selection.}
For each (model, prompt, question) cell, every trace step is scored by
gradient\,$\times$\,activation attribution toward the answer span. The scalar
target is the sum of the causal-LM next-token log probabilities of the observed
answer-span tokens, using the usual one-token shift. A single backward pass
computes the absolute value of the gradient--activation dot product at each
token and decoder layer. Per-token attributions are summed over
each step's generated token positions to give one value per
(layer, step); each layer's values are then z-scored across the trace,
and the z-scored values are averaged over layers to give a single
combined attribution score per step. Steps are ranked by descending
score, and the top-\textbf{2} steps are selected as anchors per cell
(\texttt{n\_anchors}$=2$).

\paragraph{Intervention site and backend aggregation.}
The reported mechanistic feature uses one pinned residual-stream site at
decoder layer 18 on both backends. In the PyTorch/CUDA extraction, each of
the two selected anchors is ablated in a separate redecoding and
the two correctness indicators are averaged; one matched control is ablated
separately. Thus
$A_{\mathrm{anchor}}=(I_1+I_2)/2$, $A_{\mathrm{control}}=I_c$, and
$\Delta A^{\mathrm{mech}}=I_c-(I_1+I_2)/2$, yielding support
$\{-1,-\tfrac{1}{2},0,+\tfrac{1}{2},+1\}$. The MLX extraction uses one
joint anchor-versus-control comparison at the same layer and site, yielding the coarser
support $\{-1,0,+1\}$. It records 31 non-zero values out of 1200, against 70
out of 1200 on CUDA. This is the only feature aggregated differently across
the stacks. It does not carry the cross-stack agreement reported in
Table~\ref{tab:main-hcds}, which rests on the five features measured
identically, and Appendix~\ref{app:sensitivity} shows that the GSM8K
conclusion survives dropping the mechanistic feature altogether.

\paragraph{Intervention mode (what is suppressed).}
The reported analysis uses \texttt{residual\_zero}: at layer 18, the
residual-stream output is multiplied by $0$ at the selected step's token
positions. Other token positions and layers are unchanged.
For each CUDA anchor or control ablation, the full original prefix through the
last segmented step is recomputed once with the hook active and
\texttt{use\_cache=True}, producing a fresh KV cache. The hook is then removed
and greedy decoding continues from that intervened cache. Thus the cached
prefix is recomputed rather than reused from the unperturbed run. The resulting
completion is parsed for the final answer and compared
to the unperturbed baseline; \texttt{intervention\_correct} is the
indicator that the post-intervention prediction matches gold.

\paragraph{Control-step selection.}
After anchors are chosen, the control step is selected
\emph{deterministically} as the non-anchor step nearest the anchor set
under the distance $|\rho_i - \bar{\rho}_{\mathrm{anchor}}| +
0.01\,|\ell_i - \bar{\ell}_{\mathrm{anchor}}|$, where
$\rho_i = \mathrm{idx}_i/(n-1)$ is a step's relative position in a
trace of $n$ steps and $\ell_i$ is its token length, with one control
step per cell (\texttt{n\_controls}$=1$). Controls are therefore
matched on trace position, with step length as a weak tie-break; the
selection involves no sampling and no random seed, and the control is
suppressed at the same (layer, site) pair as the anchors. Controls
are \emph{not} matched on step \emph{function} --- whether a step
performs a calculation, restates the problem, or summarises --- which
remains a limitation called out in
Section~\ref{sec:limitations}.

\section{Per-Feature Effect Sizes and Pole Distributions}
\label{app:per-feature}

Table~\ref{tab:per-feature} in Section~\ref{sec:robustness} decomposes the
length-adjusted score on GSM8K under PyTorch/CUDA. This appendix gives the
decomposition for all eight (backend, model, dataset) cells and the distributional
view collapsed by the contrast: each feature's mean z-scored value under the
three prompt conditions.

The contrast for feature $j$ on question $q$ is
$\delta_j(q) = |z_j^{\mathrm{neu}} - z_j^{\mathrm{no\text{-}cot}}|
             - |z_j^{\mathrm{neu}} - z_j^{\mathrm{cot}}|$,
computed only for questions where feature $j$ is defined at all three poles.
Five features satisfy this on almost every question; mechanistic sensitivity
does so for zero to three questions per cell, precluding effect estimation. We
report counts rather than statistics from three points.

\begin{table}[!htbp]
\centering
\footnotesize
\setlength{\tabcolsep}{4pt}
\begin{tabular}{@{}lrrrrrr@{}}
\toprule
& \multicolumn{3}{c}{Contrast} & \multicolumn{3}{c}{Mean $z$ by pole} \\
\cmidrule(lr){2-4}\cmidrule(lr){5-7}
Feature & $\bar{\delta}_j$ & $d$ & $p$ & no-CoT & CoT & neutral \\
\midrule
\multicolumn{7}{@{}l}{\emph{CUDA Thinking GSM8K}} \\
\quad latency & $-0.11$ & $-0.19$ & $0.175$ & $+0.07$ & $+0.11$ & $-0.19$ \\
\quad entropy mean & $+1.05$ & $+0.93$ & $2.7\times10^{-8}$ & $+0.14$ & $-0.09$ & $-0.05$ \\
\quad entropy slope & $+1.22$ & $+1.13$ & $2.0\times10^{-10}$ & $-0.03$ & $-0.06$ & $+0.10$ \\
\quad paraphrase consistency & $+0.67$ & $+0.66$ & $2.5\times10^{-5}$ & $-0.05$ & $-0.01$ & $+0.05$ \\
\quad perturbation sensitivity & $+0.33$ & $+0.24$ & $0.101$ & $+0.10$ & $+0.01$ & $-0.11$ \\
\quad mechanistic sensitivity & \multicolumn{6}{c}{not estimable ($n=1$)} \\
\midrule
\multicolumn{7}{@{}l}{\emph{CUDA Thinking StrategyQA}} \\
\quad latency & $+0.15$ & $+0.25$ & $0.083$ & $+0.36$ & $+0.26$ & $-0.61$ \\
\quad entropy mean & $+0.11$ & $+0.19$ & $0.194$ & $-0.29$ & $-0.31$ & $+0.60$ \\
\quad entropy slope & $+1.10$ & $+0.96$ & $1.6\times10^{-8}$ & $-0.08$ & $-0.06$ & $+0.14$ \\
\quad paraphrase consistency & $+0.16$ & $+0.13$ & $0.377$ & $-0.04$ & $+0.13$ & $-0.09$ \\
\quad perturbation sensitivity & $-0.03$ & $-0.03$ & $0.829$ & $+0.02$ & $+0.08$ & $-0.10$ \\
\quad mechanistic sensitivity & \multicolumn{6}{c}{not estimable ($n=0$)} \\
\midrule
\multicolumn{7}{@{}l}{\emph{CUDA Instruct GSM8K}} \\
\quad latency & $+0.07$ & $+0.49$ & $0.001$ & $-0.01$ & $+0.02$ & $-0.01$ \\
\quad entropy mean & $+0.70$ & $+0.57$ & $1.8\times10^{-4}$ & $+0.06$ & $+0.04$ & $-0.10$ \\
\quad entropy slope & $+0.83$ & $+0.56$ & $2.5\times10^{-4}$ & $-0.00$ & $+0.01$ & $-0.01$ \\
\quad paraphrase consistency & $+0.87$ & $+0.77$ & $1.8\times10^{-6}$ & $+0.01$ & $+0.00$ & $-0.01$ \\
\quad perturbation sensitivity & $+0.33$ & $+0.23$ & $0.114$ & $-0.02$ & $+0.01$ & $+0.01$ \\
\quad mechanistic sensitivity & \multicolumn{6}{c}{not estimable ($n=1$)} \\
\midrule
\multicolumn{7}{@{}l}{\emph{CUDA Instruct StrategyQA}} \\
\quad latency & $+0.13$ & $+0.14$ & $0.342$ & $-0.25$ & $-0.27$ & $+0.52$ \\
\quad entropy mean & $-0.00$ & $-0.00$ & $0.977$ & $-0.02$ & $-0.45$ & $+0.47$ \\
\quad entropy slope & $+0.31$ & $+0.24$ & $0.101$ & $-0.08$ & $-0.18$ & $+0.26$ \\
\quad paraphrase consistency & $-0.00$ & $-0.00$ & $0.999$ & $-0.03$ & $+0.10$ & $-0.07$ \\
\quad perturbation sensitivity & $+0.17$ & $+0.13$ & $0.378$ & $+0.02$ & $+0.13$ & $-0.15$ \\
\quad mechanistic sensitivity & \multicolumn{6}{c}{not estimable ($n=0$)} \\
\midrule
\multicolumn{7}{@{}l}{\emph{MLX Thinking GSM8K}} \\
\quad latency & $-0.12$ & $-0.10$ & $0.501$ & $+0.18$ & $+0.35$ & $-0.53$ \\
\quad entropy mean & $+0.95$ & $+0.83$ & $3.7\times10^{-7}$ & $+0.11$ & $-0.09$ & $-0.03$ \\
\quad entropy slope & $+1.19$ & $+1.02$ & $3.0\times10^{-9}$ & $-0.02$ & $-0.05$ & $+0.07$ \\
\quad paraphrase consistency & $+0.71$ & $+0.74$ & $3.4\times10^{-6}$ & $-0.02$ & $+0.05$ & $-0.03$ \\
\quad perturbation sensitivity & $+0.42$ & $+0.31$ & $0.035$ & $+0.04$ & $+0.09$ & $-0.13$ \\
\quad mechanistic sensitivity & \multicolumn{6}{c}{not estimable ($n=1$)} \\
\midrule
\multicolumn{7}{@{}l}{\emph{MLX Thinking StrategyQA}} \\
\quad latency & $-0.04$ & $-0.04$ & $0.760$ & $+0.06$ & $+0.16$ & $-0.22$ \\
\quad entropy mean & $+0.05$ & $+0.08$ & $0.586$ & $-0.25$ & $-0.41$ & $+0.66$ \\
\quad entropy slope & $+1.12$ & $+0.99$ & $7.2\times10^{-9}$ & $-0.09$ & $-0.07$ & $+0.15$ \\
\quad paraphrase consistency & $+0.25$ & $+0.22$ & $0.127$ & $-0.05$ & $+0.01$ & $+0.05$ \\
\quad perturbation sensitivity & $+0.34$ & $+0.32$ & $0.028$ & $+0.05$ & $-0.10$ & $+0.05$ \\
\quad mechanistic sensitivity & \multicolumn{6}{c}{not estimable ($n=0$)} \\
\midrule
\multicolumn{7}{@{}l}{\emph{MLX Instruct GSM8K}} \\
\quad latency & $+0.06$ & $+0.05$ & $0.705$ & $+0.00$ & $+0.08$ & $-0.08$ \\
\quad entropy mean & $+0.69$ & $+0.62$ & $6.4\times10^{-5}$ & $+0.04$ & $+0.08$ & $-0.13$ \\
\quad entropy slope & $+0.83$ & $+0.56$ & $2.7\times10^{-4}$ & $-0.01$ & $+0.01$ & $-0.00$ \\
\quad paraphrase consistency & $+0.87$ & $+0.83$ & $4.1\times10^{-7}$ & $-0.00$ & $+0.05$ & $-0.05$ \\
\quad perturbation sensitivity & $+0.42$ & $+0.32$ & $0.029$ & $-0.03$ & $-0.04$ & $+0.07$ \\
\quad mechanistic sensitivity & \multicolumn{6}{c}{not estimable ($n=3$)} \\
\midrule
\multicolumn{7}{@{}l}{\emph{MLX Instruct StrategyQA}} \\
\quad latency & $-0.06$ & $-0.06$ & $0.692$ & $+0.11$ & $+0.10$ & $-0.20$ \\
\quad entropy mean & $+0.09$ & $+0.10$ & $0.476$ & $-0.04$ & $-0.36$ & $+0.40$ \\
\quad entropy slope & $+0.30$ & $+0.24$ & $0.099$ & $-0.09$ & $-0.17$ & $+0.26$ \\
\quad paraphrase consistency & $-0.07$ & $-0.06$ & $0.662$ & $-0.08$ & $-0.02$ & $+0.10$ \\
\quad perturbation sensitivity & $-0.20$ & $-0.15$ & $0.286$ & $+0.01$ & $+0.01$ & $-0.02$ \\
\quad mechanistic sensitivity & \multicolumn{6}{c}{not estimable ($n=0$)} \\
\bottomrule
\end{tabular}
\caption{Per-feature contrast and per-pole means, length-adjusted, $n{=}50$ per
cell. Positive $\bar{\delta}_j$ means neutral behavior sits nearer the
explicit-CoT pole on that feature alone. Pole columns are mean z-scored feature
values under each prompt condition, z-scored within (model, dataset).}
\label{tab:per-feature-full}
\end{table}

\paragraph{What the pole columns show.} The contrast is a statement about
per-question distances and can be positive even where the pole means are close,
which is the usual case here: the effect lives in the within-question
comparison, not in a shift of the marginal distribution. StrategyQA is the
visible exception --- there the neutral pole separates from both explicit poles
on latency and entropy mean by $0.3$ to $1.1$ z-units, in the direction of
longer, higher-entropy neutral outputs, while the per-question contrast on those
same features is small and mostly non-significant. That combination is what a
condition-level shift without a per-question effect looks like, and it is
consistent with the reading in Appendix~\ref{app:strategyqa}: on a boolean task
the no-CoT pole never becomes answer-only, so the contrast that HCDS is built
from is compressed even when the conditions differ on average.

\section{Sensitivity to Feature-Subset Weighting}
\label{app:sensitivity}

Because HCDS combines six features with equal weighting in z-scored
space, a natural reviewer concern is whether the headline conclusion
depends on that weighting choice. We address this with a discrete
sensitivity analysis: for each (dataset, model) we recompute HCDS
on every $2^k - 1$ non-empty subset of the $k$ recorded HCDS features
in that dataset, and record the fraction of subsets
for which HCDS is positive and significantly positive ($p<0.05$
two-sided one-sample $t$-test).

\begin{table}[!htbp]
\centering
\small
\setlength{\tabcolsep}{5pt}
\begin{tabular}{@{}lllrrr@{}}
\toprule
Backend & Dataset & Model & Subsets & \% positive & \% sig.\ positive \\
\midrule
PyTorch/CUDA & GSM8K $n=50$       & Instruct & 63 & 96.8\% & 84.1\% \\
PyTorch/CUDA & GSM8K $n=50$       & Thinking & 63 & 95.2\% & 87.3\% \\
PyTorch/CUDA & StrategyQA $n=50$  & Instruct & 63 & 88.9\% & 0.0\% \\
PyTorch/CUDA & StrategyQA $n=50$  & Thinking & 63 & 88.9\% & 52.4\% \\
\midrule
Apple MLX    & GSM8K $n=50$       & Instruct & 63 & 98.4\% & 88.9\% \\
Apple MLX    & GSM8K $n=50$       & Thinking & 63 & 95.2\% & 90.5\% \\
Apple MLX    & StrategyQA $n=50$  & Instruct & 63 & 28.6\% & 0.0\% \\
Apple MLX    & StrategyQA $n=50$  & Thinking & 63 & 92.1\% & 57.1\% \\
\bottomrule
\end{tabular}
\caption{Length-adjusted HCDS sign and significance coverage across all $2^6-1=63$ non-empty feature subsets per (backend, dataset, model). On GSM8K, the length-adjusted score is positive in 95.2\%--98.4\% and significantly positive in 84.1\%--90.5\% of all subsets across both models and backends. On StrategyQA, Instruct is significant in 0\% of subsets (consistent with the chance-accuracy floor in Appendix~\ref{app:strategyqa}), while Thinking survives in 52.4\%--57.1\%.}
\label{tab:sensitivity}
\end{table}

\paragraph{Where the negative-sign subsets fall.} GSM8K's only negative subsets
are degenerate combinations of latency and/or mechanistic sensitivity without
behavioral or linguistic features. Every subset with a linguistic or entropy
feature remains positive, consistent with Section~\ref{sec:robustness}.

\begin{figure}[!htbp]
\refstepcounter{sensitivityfigure}
\label{fig:ablation-mlx}
\centering
\includegraphics[width=\columnwidth,trim=0 18bp 0 170.4bp,clip]{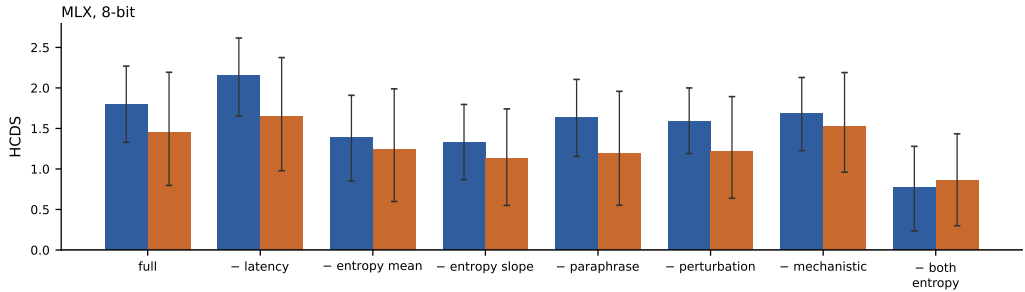}
\caption*{\textbf{Figure~\thesensitivityfigure: MLX feature-ablation robustness.}
The MLX 8-bit arm of the feature-ablation analysis on GSM8K, $n{=}50$, across
both models. The pattern matches the PyTorch/CUDA bf16 arm in
Figure~\ref{fig:ablation}: removing latency raises HCDS, while removing both
entropy features produces the largest reduction. Error bars show 95\% bootstrap
CIs.}
\end{figure}

The MLX arm of the leave-one-feature-out analysis reproduces the CUDA pattern
(Figure~\ref{fig:ablation-mlx}).

\paragraph{Implications.} The headline conclusion---positive HCDS aligns neutral
behavior with explicit CoT---holds for the vast majority of GSM8K feature weights. Equal weighting
is therefore not load-bearing; almost any non-trivial weighting containing
behavioral or uncertainty features preserves the sign.

\paragraph{Continuous weight sampling.} To test robustness to arbitrary continuous weightings, we draw $10{,}000$ weight vectors uniformly from the feature simplex ($\mathrm{Dirichlet}(1,\dots,1)$) and recompute HCDS under the weighted metric $D_w(a,b)=\sqrt{\sum_j w_j (a_j-b_j)^2}$, reproducing the length-adjusted GSM8K baselines in Table~\ref{tab:main-hcds}. Instruct HCDS remains positive in $>98\%$ of sampled weightings and significantly positive in $>90\%$; Thinking remains positive in $>94\%$ and significantly positive in $>84\%$. We deliberately do not \emph{optimize} the weights: HCDS uses no
per-question hidden-CoT label, so tuning weights to enlarge the
effect would be circular; equal weighting is retained as a principled
default whose robustness we verify rather than fit.

\section{Multiple-Comparisons Correction}
\label{app:multiplicity}

Every reported HCDS has an uncorrected two-sided one-sample $t$-test, including sixteen
length-adjusted cells, twenty-four leave-one-out ablations, and a
sixty-three-subset sweep per cell. We therefore report raw and Holm--Bonferroni
(family-wise error rate) and Benjamini--Hochberg (false discovery rate) adjusted
p-values. Families group tests by their claims, not by convenience.

\paragraph{Primary family.} The sixteen length-adjusted cells --- two backends
$\times$ two models $\times$ four datasets --- carry the paper's confirmatory
claims and are corrected together (Table~\ref{tab:multiplicity}). Correcting
negative controls alongside positive cells controls false positives but does not
establish practical equivalence to zero; it only makes positive declarations harder.

Seven of sixteen cells are significant at uncorrected $\alpha=0.05$; six survive
Holm. All four headline GSM8K cells survive comfortably
(largest adjusted $p = 2.5\times10^{-3}$), as do both Thinking StrategyQA cells.
The single cell that loses significance is the CUDA Instruct arithmetic
\emph{negative control}, the marginal exception discussed in
Section~\ref{sec:discussion} (Table~\ref{tab:multiplicity}). Under Holm, no
length-adjusted negative-control cell is significantly positive. This controls the
reported false-positive rate but does not establish equivalence to zero.

\begin{table}[!htbp]
\centering
\small
\setlength{\tabcolsep}{5pt}
\begin{tabular}{@{}lllrrrrc@{}}
\toprule
Backend & Model & Dataset & HCDS & $p$ (raw) & $p$ (Holm) & $p$ (BH) & Sig.\ (Holm) \\
\midrule
PyTorch/CUDA & Thinking & GSM8K & $+1.872$ & $1.2\times10^{-7}$ & $1.8\times10^{-6}$ & $9.5\times10^{-7}$ & Yes \\
PyTorch/CUDA & Thinking & StrategyQA & $+0.724$ & $3.8\times10^{-4}$ & $0.005$ & $0.001$ & Yes \\
PyTorch/CUDA & Thinking & Arithmetic$^{\dagger}$ & $+0.392$ & $0.179$ & $1.000$ & $0.286$ & No \\
PyTorch/CUDA & Thinking & Factual$^{\dagger}$ & $+0.131$ & $0.586$ & $1.000$ & $0.648$ & No \\
PyTorch/CUDA & Instruct & GSM8K & $+1.410$ & $1.9\times10^{-4}$ & $0.003$ & $7.8\times10^{-4}$ & Yes \\
PyTorch/CUDA & Instruct & StrategyQA & $+0.291$ & $0.283$ & $1.000$ & $0.377$ & No \\
PyTorch/CUDA & Instruct & Arithmetic$^{\dagger}$ & $+0.578$ & $0.046$ & $0.461$ & $0.105$ & No \\
PyTorch/CUDA & Instruct & Factual$^{\dagger}$ & $+0.267$ & $0.088$ & $0.794$ & $0.176$ & No \\
\midrule
Apple MLX & Thinking & GSM8K & $+1.799$ & $1.3\times10^{-9}$ & $2.0\times10^{-8}$ & $2.0\times10^{-8}$ & Yes \\
Apple MLX & Thinking & StrategyQA & $+0.738$ & $0.001$ & $0.013$ & $0.003$ & Yes \\
Apple MLX & Thinking & Arithmetic$^{\dagger}$ & $+0.323$ & $0.225$ & $1.000$ & $0.327$ & No \\
Apple MLX & Thinking & Factual$^{\dagger}$ & $+0.334$ & $0.101$ & $0.808$ & $0.180$ & No \\
Apple MLX & Instruct & GSM8K & $+1.449$ & $1.7\times10^{-4}$ & $0.002$ & $7.8\times10^{-4}$ & Yes \\
Apple MLX & Instruct & StrategyQA & $-0.149$ & $0.544$ & $1.000$ & $0.648$ & No \\
Apple MLX & Instruct & Arithmetic$^{\dagger}$ & $+0.173$ & $0.608$ & $1.000$ & $0.648$ & No \\
Apple MLX & Instruct & Factual$^{\dagger}$ & $+0.050$ & $0.803$ & $1.000$ & $0.803$ & No \\
\bottomrule
\end{tabular}
\caption{Holm--Bonferroni and Benjamini--Hochberg adjusted p-values across the
sixteen length-adjusted HCDS cells, corrected as one family. $^{\dagger}$ marks
negative-control tasks, where the expected result is near zero. Six of the seven
uncorrected significant cells survive Holm; the one that does not is a negative
control. Non-significance after correction is not an equivalence result.}
\label{tab:multiplicity}
\end{table}

\paragraph{Robustness families.} The twenty-four leave-one-out ablations behind
Figures~\ref{fig:ablation} and~\ref{fig:ablation-mlx} are robustness probes
around an already-tested claim, so they form their own family; all twenty-four
survive Holm at $\alpha=0.05$ (largest adjusted $p = 4.0\times10^{-3}$). The
subset sweep behind
Table~\ref{tab:sensitivity} reports a \emph{fraction} of significant subsets, so
we recompute that fraction after correcting within each sweep. It falls, as it
must when sixty-three tests are corrected together: on GSM8K from
$84.1\%$--$90.5\%$ raw to $65.1\%$--$84.1\%$ under Holm and $84.1\%$--$90.5\%$
under BH. The sign-based claim that appendix actually makes --- that HCDS is
positive under almost any non-trivial weighting --- is unaffected, since
correction changes no point estimate.

\paragraph{Scope.} These corrections address multiplicity across reported tests.
They do not address the deeper issue raised in Section~\ref{sec:limitations}:
with deterministic decoding and one trial per question, the unit of variation is
question identity rather than model stochasticity, and no p-value adjustment
changes what the test is a test of.

\section{Correlation-Aware (Mahalanobis) HCDS}
\label{app:mahalanobis}

Appendix~\ref{app:sensitivity} establishes that the headline result does not
depend on how the six features are \emph{weighted}. A distinct assumption
survives that analysis: Euclidean distance in z-scored space treats the six
features as mutually independent. They are not necessarily so --- all six are
computed from the same generated response, and several are statistics over
generated tokens, which is precisely the coupling that motivates the length
adjustment in Section~\ref{subsec:length-adjust}. Where two features are
correlated, a Euclidean contrast counts their shared component twice, so a
prompt condition that moves one correlated block of features is weighted more
heavily than one that moves an equal number of independent features.

Mahalanobis distance removes that double-counting. We replace $D$ with
\[
\begin{aligned}
  D_{\mathrm{M},\mathcal{J}_q}(a,b)
  &= \sqrt{(a_{\mathcal{J}_q}-b_{\mathcal{J}_q})^{\top}
  (S_{\lambda})_{\mathcal{J}_q}^{-1}
  (a_{\mathcal{J}_q}-b_{\mathcal{J}_q})},\\
  D_{\mathrm{M}}(a,b)
  &= \sqrt{\frac{|\mathcal{F}|}{|\mathcal{J}_q|}}\,
  D_{\mathrm{M},\mathcal{J}_q}(a,b),
  \qquad S_{\lambda} = (1-\lambda)\,S + \lambda I.
\end{aligned}
\]
where $S$ is the feature covariance estimated within the same $(m,d)$ group used
for z-scoring. Because the inputs are already z-scored, $S$ is the feature
\emph{correlation} matrix, which makes the comparison unusually clean:
$\lambda = 1$ gives $S_{\lambda} = I$ and recovers the published Euclidean score
\emph{exactly}, with bit-identical agreement across all sixteen cells (eight
length-residualised and eight unadjusted),
while $\lambda = 0$ is unregularised whitening. Any difference between the two
is therefore attributable to feature correlation alone, and not to a change of
scale, weighting, or estimator. Shrinkage is necessary because $S$ is estimated
from $150$ rows over six features and $\Delta A^{\mathrm{mech}}$ is defined in
only ${\sim}40\%$ of cells; we set $\lambda$ by Ledoit-Wolf and report the value
used. The same three NaN policies apply, with distances over a subset of
features taken against the corresponding submatrix of $S_{\lambda}$: the matrix
is shrunk at full rank, then subset to the features available for that distance,
and the resulting scalar distance is rescaled.

\begin{table}[!htbp]
\centering
\small
\setlength{\tabcolsep}{5pt}
\begin{tabular}{@{}lllrrrr@{}}
\toprule
Backend & Model & Task & Euclidean & Mahalanobis & $\hat{\lambda}$ & $p$ (Mahal.) \\
\midrule
PyTorch/CUDA & Thinking & GSM8K      & $+1.872$ & $+1.872$ & 1.00 & $1.2\times10^{-7}$ \\
Apple MLX    & Thinking & GSM8K      & $+1.799$ & $+1.738$ & 0.73 & $1.3\times10^{-9}$ \\
PyTorch/CUDA & Instruct & GSM8K      & $+1.410$ & $+1.316$ & 0.38 & $3.0\times10^{-4}$ \\
Apple MLX    & Instruct & GSM8K      & $+1.449$ & $+1.372$ & 0.61 & $3.1\times10^{-4}$ \\
\midrule
PyTorch/CUDA & Thinking & StrategyQA & $+0.724$ & $+0.716$ & 0.76 & $4.0\times10^{-4}$ \\
Apple MLX    & Thinking & StrategyQA & $+0.738$ & $+0.711$ & 0.81 & $1.7\times10^{-3}$ \\
PyTorch/CUDA & Instruct & StrategyQA & $+0.291$ & $+0.285$ & 0.70 & $0.29$ \\
Apple MLX    & Instruct & StrategyQA & $-0.149$ & $-0.147$ & 0.79 & $0.55$ \\
\midrule
PyTorch/CUDA & Thinking & Arithmetic & $+0.392$ & $+0.331$ & 0.31 & $0.26$ \\
Apple MLX    & Thinking & Arithmetic & $+0.323$ & $+0.293$ & 0.35 & $0.27$ \\
PyTorch/CUDA & Instruct & Arithmetic & $+0.578$ & $+0.418$ & 0.27 & $0.14$ \\
Apple MLX    & Instruct & Arithmetic & $+0.173$ & $+0.130$ & 0.77 & $0.70$ \\
PyTorch/CUDA & Thinking & Factual    & $+0.131$ & $+0.131$ & 1.00 & $0.59$ \\
Apple MLX    & Thinking & Factual    & $+0.334$ & $+0.386$ & 0.56 & $0.058$ \\
PyTorch/CUDA & Instruct & Factual    & $+0.267$ & $+0.246$ & 0.48 & $0.12$ \\
Apple MLX    & Instruct & Factual    & $+0.050$ & $+0.066$ & 0.55 & $0.74$ \\
\bottomrule
\end{tabular}
\caption{Length-adjusted HCDS under Euclidean and Mahalanobis distance, $n=50$
per cell, both backends. $\hat{\lambda}$ is the Ledoit-Wolf shrinkage intensity
toward the identity; $\hat{\lambda}=1$ means the data supply no evidence of
exploitable correlation structure and the two metrics coincide by construction.
GSM8K remains significant in all four cells; no negative-control cell is
significantly positive under Mahalanobis distance.}
\label{tab:mahalanobis}
\end{table}

\paragraph{GSM8K survives whitening.} The reasoning signal is not an artifact of
correlated features. Under Ledoit-Wolf shrinkage the four GSM8K cells move by at
most $0.09$ and remain significant at $p<10^{-3}$ or better. Under fully
unregularised whitening ($\lambda=0$) they read $+1.747$ and $+1.713$ (Thinking,
CUDA and MLX) and $+1.268$ and $+1.324$ (Instruct), i.e.\ a reduction of at most
$0.13$ from the published values. The path from $\lambda=0$ to $\lambda=1$ is
shallow and monotone in every cell, with no threshold at which the result
appears or disappears.

\paragraph{The one marginal negative control resolves.} The marginal CUDA
Instruct arithmetic control in Table~\ref{tab:negcontrol} is no longer
significantly positive once features are decorrelated (Table~\ref{tab:mahalanobis}).
Under Mahalanobis distance no control cell is significantly positive. We report this as a consistency
check rather than as grounds for promoting the correlation-aware variant: one
cell moving across the significance threshold at $n=50$ is weak evidence on its
own, and the movement is not uniformly in the metric's favour --- the MLX
Thinking factual-lookup control remains non-significant but moves closer to the threshold.

\paragraph{Length residualisation has already done most of the decorrelation.}
The two metrics agree closely on adjusted features (mean $|r| = 0.105$ for
Thinking and $0.158$ for Instruct on CUDA GSM8K; maximum absolute displacement
across all sixteen cells $0.16$), but diverge substantially on unadjusted
features, where whitening moves StrategyQA by $-0.506$ (Thinking) and $-0.561$
(Instruct). Output length was the dominant source of shared variance among the
features, so residualising on $\log n^{\mathrm{gen}}$ removes most of the
correlation structure as a side effect. This is an independent argument for the
length adjustment: it was introduced to reduce positive calibration-control scores, and it
additionally makes the choice of metric close to immaterial. The largest
correlation surviving adjustment is between perturbation sensitivity and
mechanistic sensitivity ($r=0.54$ on CUDA Instruct GSM8K), which is expected ---
both are intervention-response features --- and is the main reason the Instruct
cells shrink more than the Thinking cells.

\paragraph{Which to report.} We retain the Euclidean score as primary. It is the
simpler and less parameterised of the two, it requires no covariance estimate
from $50$ questions, and the correlation-aware variant reproduces it. The
Mahalanobis result is offered as evidence that equal weighting \emph{and}
implicit independence are both non-load-bearing, closing the assumption left
open by Appendix~\ref{app:sensitivity}.

\end{document}